\pdfoutput=1
\documentclass[11pt]{article}

\usepackage[final]{acl}

\usepackage{times}
\usepackage{latexsym}
\usepackage{xcolor}
\usepackage{pgf}
\usepackage{amsmath}

\usepackage{xcolor}
\usepackage{colortbl}
\usepackage{pgf}
\usepackage{booktabs}      % Для красивых линий
\usepackage{pgfplots}      % Для использования \pgfmathparse и вычислений
\pgfplotsset{compat=1.17}  % Совместимость с конкретной версией

\usepackage{times}
\usepackage{latexsym}
\usepackage{graphicx}
\usepackage{amssymb}
\usepackage{stfloats}

\usepackage{xcolor}    % Для цветов
\usepackage{colortbl}  % Чтобы закрашивать ячейки
\usepackage{pgf}       % Чтобы делать вычисления для градиентов

\usepackage[T1]{fontenc}
\usepackage[utf8]{inputenc}

\usepackage{microtype}

\usepackage{inconsolata}

\usepackage{graphicx}
\usepackage{booktabs} 

\title{LLM-Microscope: Uncovering the Hidden Role of Punctuation\\
in Context Memory of Transformers}

\author{
Anton Razzhigaev\textsuperscript{1,2},  % $\diamondsuit$}, 
Matvey Mikhalchuk\textsuperscript{1,4}, % $\diamondsuit$},
Temurbek Rahmatullaev\textsuperscript{1,3,4}, \\% $\diamondsuit$},
\bf Elizaveta Goncharova\textsuperscript{1,3}, 
Polina Druzhinina\textsuperscript{1,2}, \\
\bf Ivan Oseledets\textsuperscript{1,2}, and
Andrey Kuznetsov\textsuperscript{1}\\
\textsuperscript{1}AIRI, 
\textsuperscript{2}Skoltech,
\textsuperscript{3}HSE University,\\
\textsuperscript{4}Lomonosov Moscow State University\\
\href{mailto:razzhigaev@skol.tech}{razzhigaev@skol.tech} \\ % 
}

\begin{document}
\maketitle

\begin{abstract} We introduce methods to quantify how Large Language Models (LLMs) encode and store contextual information, revealing that tokens often seen as minor (e.g., determiners, punctuation) carry surprisingly high context. Notably, removing these tokens — especially stopwords, articles, and commas — consistently degrades performance on MMLU and BABILong-4k, even if removing only irrelevant tokens. Our analysis also shows a strong correlation between contextualization and \emph{linearity}, where linearity measures how closely the transformation from one layer’s embeddings to the next can be approximated by a single linear mapping. These findings underscore the hidden importance of ``filler'' tokens in maintaining context. For further exploration, we present \textbf{LLM-Microscope}, an open-source toolkit that assesses token-level nonlinearity, evaluates contextual memory, visualizes intermediate layer contributions (via an adapted Logit Lens), and measures the intrinsic dimensionality of representations. This toolkit illuminates how seemingly trivial tokens can be critical for long-range understanding\footnote{\url{https://github.com/AIRI-Institute/LLM-Microscope/tree/main}}.\end{abstract}

\section{Introduction}

Large Language Models (LLMs) have significantly advanced the field of natural language processing, achieving remarkable results across a wide range of tasks. Despite their success, the internal mechanisms by which these models operate remain largely opaque, making it challenging to interpret how they process and utilize contextual information. This opacity limits our ability to enhance model performance and to understand the reasoning behind their predictions. While recent studies have begun to uncover specific patterns and mechanisms within LLMs \cite{wang2022interpretabilitywildcircuitindirect}, many fundamental aspects — such as their handling of step-by-step reasoning and long-range dependencies — are still not well understood. This gap in understanding hinders the development of more interpretable and efficient language models.

\begin{figure}
  \centering
   \includegraphics[width=0.9\linewidth]{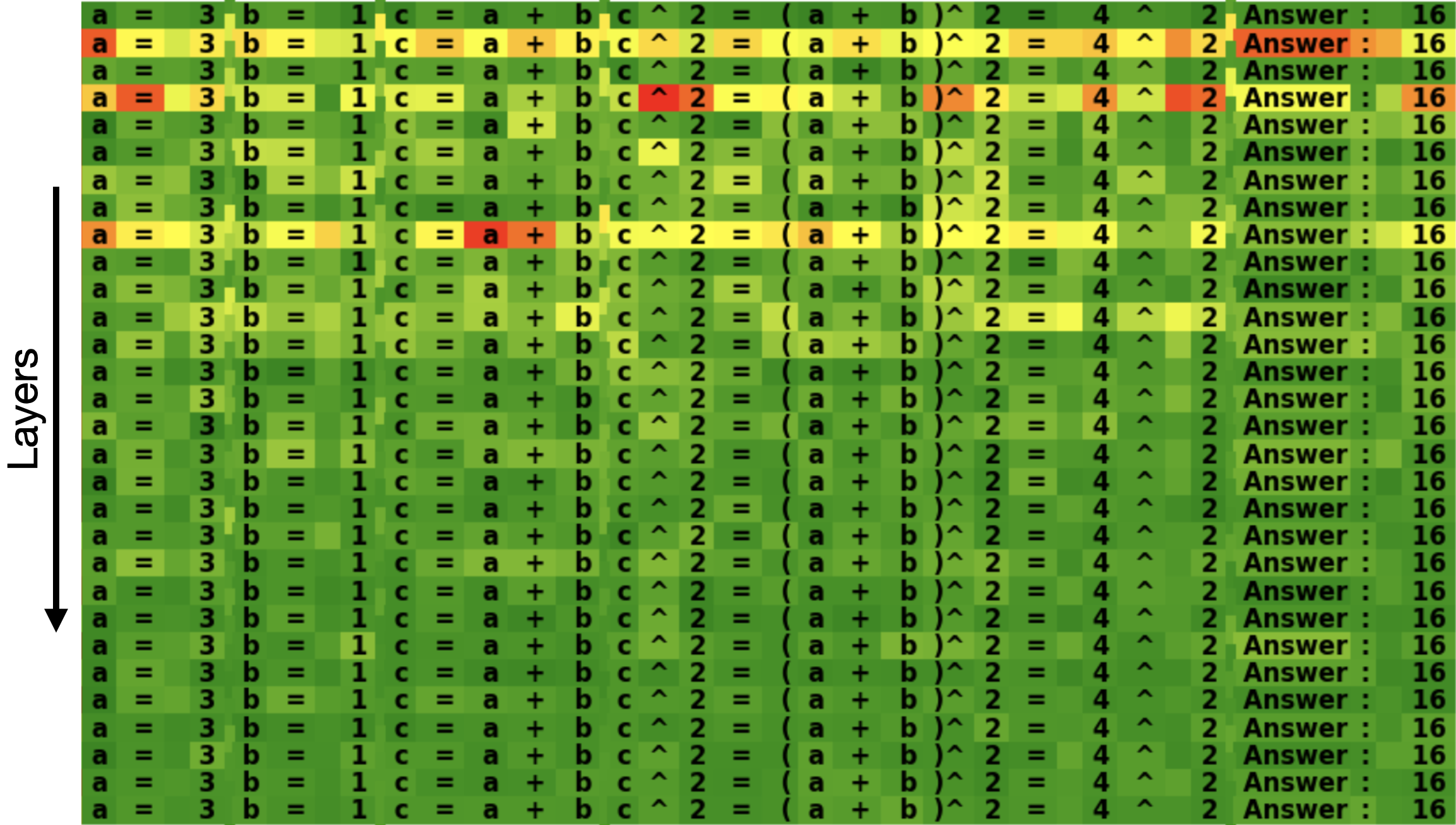}
   \caption{An example of token-wise non-linearity visualization for Llama3-8B.}
   \label{fig:math}
\end{figure}

To bridge this gap, we introduce \textbf{LLM-Microscope}, a comprehensive framework designed to analyze and visualize the internal behaviors of LLMs. Our toolkit offers a suite of methods that enable researchers to inspect how models encode and aggregate contextual information:

\begin{itemize} 
\item \textbf{Contextualization assessment}: We present a method for measuring contextualization, allowing the identification of tokens that carry the most contextual information.

\item \textbf{Token-level nonlinearity}: We measure the nonlinearity at the token level, quantifying how closely transformations between layers can be approximated by a single linear mapping.

\item \textbf{Intermediate layer analysis}: We examine how next-token prediction evolves across different layers, adapting the Logit Lens technique for multimodal LLMs. \end{itemize}

Applying these tools to various scenarios — ranging from multilingual prompts to knowledge-intensive tasks — we uncover intriguing patterns in how LLMs process and transform information. Notably, our analysis reveals that certain “filler” tokens, such as punctuation marks, stopwords, and articles, are highly contextualized and act as key aggregators in language understanding. We also find a strong correlation between linearity and contextualization scores in token representations.

Furthermore, we demonstrate the practical implications of our findings by showing that removing these tokens degrades performance on tasks requiring specialized knowledge and longer-context reasoning, such as MMLU and BABILong-4k. This performance drop persists even when we carefully remove only tokens deemed irrelevant by a strong language model (GPT-4o). These results highlight the hidden importance of seemingly “trivial” tokens in maintaining coherent context.

\textbf{LLM-Microscope} is designed to be accessible for both researchers and practitioners, providing an intuitive interface for in-depth model analysis. We offer:

\begin{itemize} \item An open-source Python package\footnote{\url{https://github.com/AIRI-Institute/LLM-Microscope/tree/main}} \item A demo website on Hugging Face Spaces\footnote{\url{https://huggingface.co/spaces/AIRI-Institute/LLM-Microscope}}\footnote{\url{https://huggingface.co/spaces/matveymih/LLM-Microscope}}\end{itemize}

\section{Related works}

\paragraph{Interpretability} There are several significant paradigms for model interpretation, each with its own distinct properties. Probing methods are designed to train classifiers based on hidden representations that are challenged in encoding specific knowledge \cite{ettinger-etal-2016-probing,belinkov-etal-2017-neural,conneau-etal-2018-cram,belinkov2022probing}. While these approaches show whether specific language features are incorporated into LLMs, they do not analyze internal representations during knowledge activation, leaving the model's behavior largely a black box.

In contrast, mechanistic interpretability introduces approaches to explore the inner behavior of models.  \citet{calderon2024behalf} mention that mechanistic interpretability aims to explore the internal representations of deep learning models through the activations of specific neurons and layer connections. A significant branch of research dedicated to examining model responses involves probing changes in behavior resulting from perturbations, noise in embeddings, or masking of network weights \cite{dai-etal-2022-knowledge,meng2022locating,olsson2022context,wang2022interpretabilitywildcircuitindirect,conmy2023}. 

Discovering interpretable features through training sparse autoencoders (SAEs) has become a promising direction in the LLM interpretation \cite{cunningham2023sparseautoencodershighlyinterpretable,yu-etal-2023-characterizing}. Typically, SAEs focus on activations of specific LLM components, such as attention heads or multilayer perceptrons (MLPs). By decomposing model computations into understandable circuits, we can see how information heads, relation heads, and MLPs encode knowledge\cite{yao2024knowledgecircuitspretrainedtransformers}.

While most research has concentrated on the analysis of Small Language Models, such as GPT-2 \cite{radford2019language} and TinyLLAMA \cite{zhang2024tinyllama}, recent work has advanced this area by proposing modifications to improve the scalability and sparsity of autoencoders for larger LLMs, such as GPT-4 or Claude 3 Sonet \cite{gao2024,templeton2024scaling}.

\paragraph{Linearity of LLM hidden states}

The study of the internal structure of transformer-based models has been of great interest among researchers \cite{nostalgebraist2020,xu-etal-2021-probing,belrose2023eliciting,din2023jump,razzhigaev-etal-2024-shape}. Several studies, such as ``Logit Lens''\footnote{\url{https://www.lesswrong.com/posts/AcKRB8wDpdaN6v6ru/interpreting-gpt-the-logit-lens}}, have explored projecting representations from the intermediate layers into the vocabulary space by observing their evolution across different layers \cite{nostalgebraist2020,belrose2023eliciting}. Relying on this research, the authors also investigate the complex structure of hidden representations through linearization \cite{anthropic2021,razzhigaev2024transformersecretlylinear}.

\paragraph{Contextualization of LLM hidden states}

One of the areas of research into the internal representations of Transformers is the embeddings contextualization analysis. Recent studies have demonstrated that sentence representations provided by Transformer decoders can contain information about the entire previous context \cite{li-etal-2023-sentence,wan2024informationleakageembeddinglarge}. \citet{wan2024informationleakageembeddinglarge} proposed two initial methods for reconstructing original texts from model’s hidden states, finding these methods effective for the embeddings from shallow layers but less effective for deeper layers, known as ``Embed Parrot.''

Our work proposes a unified framework for LLM interpretability by exploring properties such as linearity, anisotropy, and intrinsic dimension of hidden representations. We introduce new approaches to assess contextual memory in token representations and analyze intermediate layer contributions to token prediction.

\section{LLM-Microscope}

\begin{figure*}[h]
  \centering
   \includegraphics[width=0.7\linewidth]{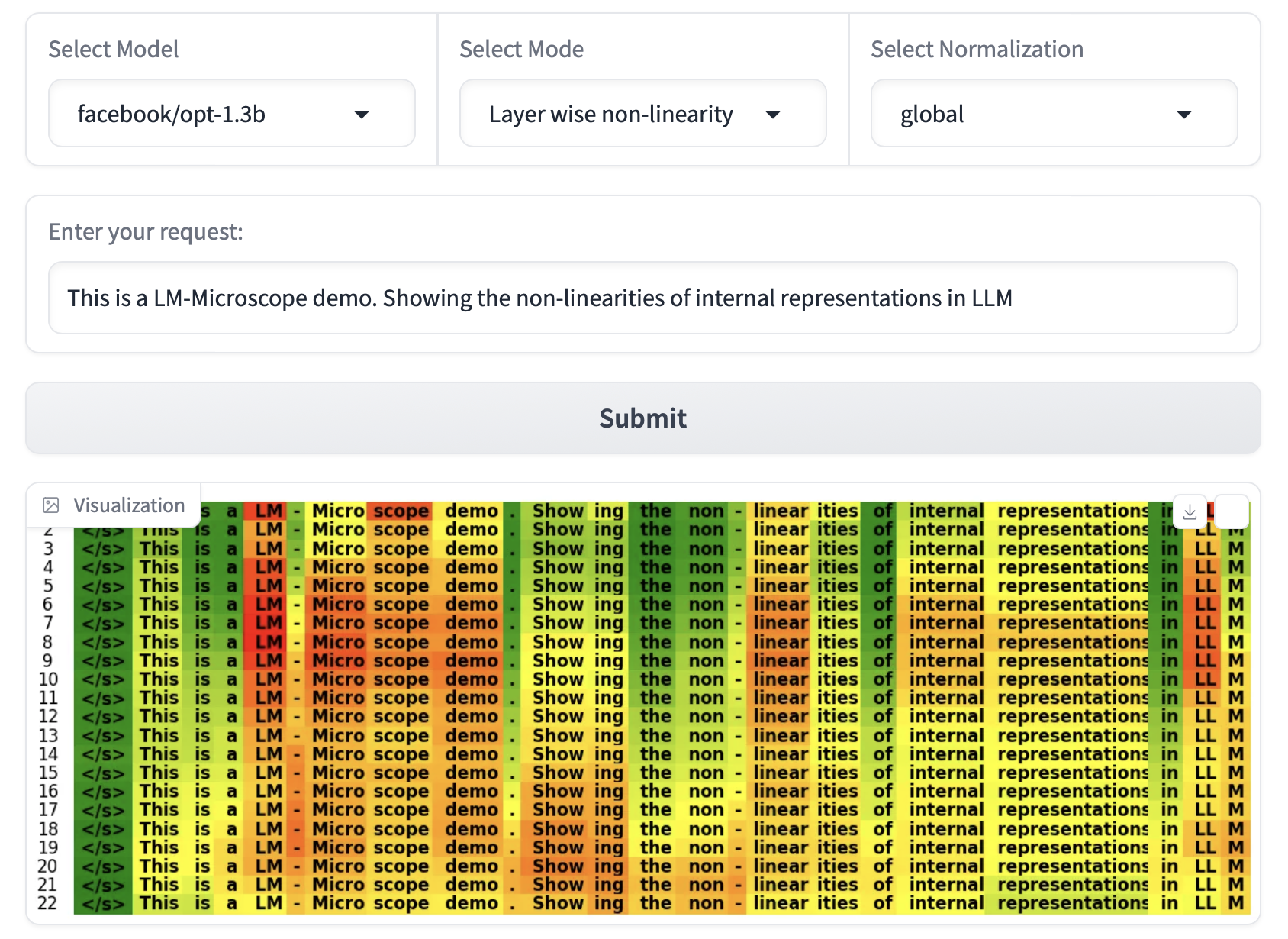}

   \caption{Interface LLM-Microscope demo system.}
   \label{fig:demo_interface}
\end{figure*}

LLM-Microscope is a framework to analyze Large Language Models' internal processes. To facilitate interactive exploration of our analysis methods, we have developed a demo system using Gradio, hosted on Hugging Face. This interface allows researchers and practitioners to apply LLM-Microscope's tools to various models and input texts in real-time.
The demo system features:
\begin{itemize}
\item Model selection: Users can choose from a variety of pre-loaded language models.
\item Text input: A text area for entering custom prompts or sentences for analysis.
\item Visualization dashboard: Upon submission, the system generates and displays:
\begin{itemize}
\item A heatmap of token-level nonlinearity across all layers
\item A line graph showing average linearity scores per layer
\item A heatmap of layer-wise contribution to final token prediction
\item A heatmap showing the contextualization level of each token
\item Visualization of the logit lens showing the preliminary predictions of the intermediate layers
\end{itemize}
\end{itemize}
The interface of our system can be found in the Figure~\ref{fig:demo_interface}.

For example, in Figure~\ref{fig:math}, one can observe patterns of nonlinearity across layers for a logical reasoning task. Different colors indicate different degrees of nonlinearity, potentially corresponding to key points in the model's reasoning process.

For users requiring more in-depth analysis or wishing to examine models not integrated into the demo, we have published our entire codebase.

\subsection{Measuring Token-level Nonlinearity}

Following the methodology for quantifying the degree of nonlinearity in token representations across model layers \cite{razzhigaev2024transformersecretlylinear} , we apply a generalized Procrustes analysis for arbitrary linear transformations. For each pair of adjacent layers $l$ and $l+1$, we compute:
\begin{equation}
A^* = \min_{A \in \mathbb{R}^{d \times d}} |\hat{H}^l A - \hat{H}^{l+1}|_F^2
\end{equation}
where $A^*$ is the optimal linear transformation found during the linearity score computation, $\hat{H}^l$ and $\hat{H}^{l+1}$ are normalized and centered matrices of token embeddings from layers $l$ and $l+1$ respectively, and $|\cdot|_F$ denotes the Frobenius norm.
The linear approximation error (nonlinearity score) for each token $i$ at layer $l$ is then calculated as:
\begin{equation}
E_i^l = |A^* h_i^l - h_i^{l+1}|_2
\end{equation}
where $h_i^l$ is the embedding of token $i$ at layer $l$.

\subsection{Assessing Contextual Memory in Token-Level Representations}
\label{contextuality}

To quantify the amount of contextual information stored in token-level representations, we propose a simple technique that uses the model's ability to reconstruct prefix information from individual token representations. This approach provides insight into how different tokens encode and preserve context across all layers of the model.

\begin{figure*}[h]
  \centering
   \includegraphics[width=0.8\linewidth]{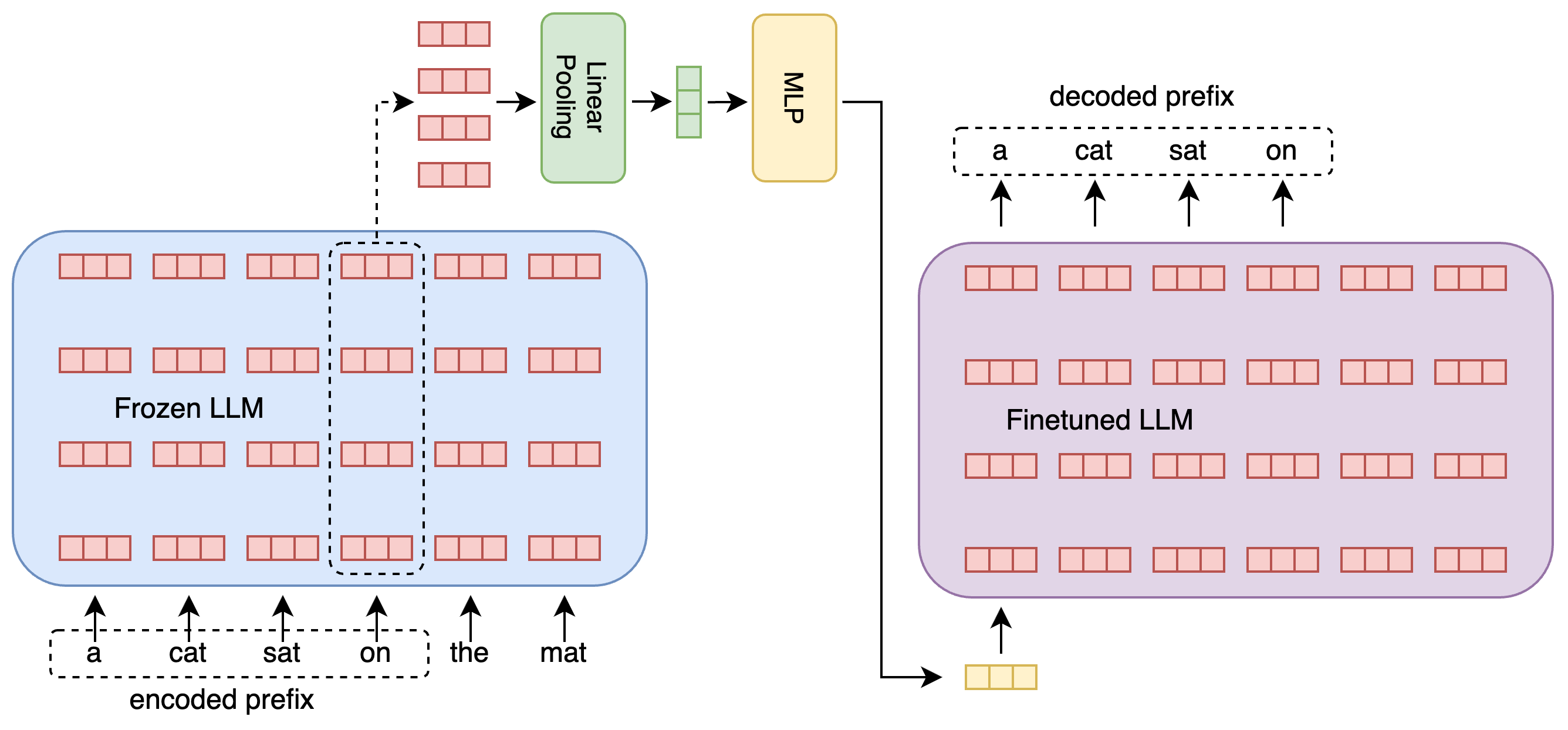}

   \caption{Prefix decoding pipeline as a contextualization assessment.}
   \label{fig:decoder_pipeline}
\end{figure*}

Our method (Figure~\ref{fig:decoder_pipeline}) consists of the following steps:

\begin{enumerate}
    \item We first process an input sequence through the examined language model, collecting hidden states for each token across all layers.
    
    \item We use a trainable linear pooling layer to combine these layer-wise embeddings into a single representation. This pooling layer is followed by a two-layer MLP.
    
    \item The resulting embedding is then used as input to a trainable copy of the original model, which attempts to reconstruct the prefix leading to the chosen token.

    \item The described system is trained with a Cross-Entropy loss to reconstruct random text fragments. For training we use the following datasets: TinyStories \cite{tinyStories}, Tiny-Textbooks\footnote{\url{https://huggingface.co/datasets/nampdn-ai/tiny-textbooks}}, Tiny-Lessons\footnote{\url{https://huggingface.co/datasets/nampdn-ai/tiny-lessons}}, Tiny-Orca-Textbooks\footnote{\url{https://huggingface.co/datasets/nampdn-ai/tiny-orca-textbooks}}, Tiny-Codes\footnote{\url{https://huggingface.co/datasets/nampdn-ai/tiny-codes}}, textbooks-are-all-you-need-lite\footnote{\url{https://huggingface.co/datasets/SciPhi/textbooks-are-all-you-need-lite}}.
    
    \item We evaluate the effectiveness of this reconstruction by computing the perplexity of the generated prefix compared to the original input.
    
\end{enumerate}

The full pipeline is depicted in the Figure~\ref{fig:decoder_pipeline}. The CrossEntropy reconstruction loss score serves as our measure of contextualization. A lower loss indicates that the token's representation contains more information about its context, as the model is able to reconstruct the previous text more accurately.

Formally, let $h_i^l$ denote the hidden state of the $i$-th token at layer $l$. Our pooling function $f$ and subsequent MLP $g$ can be expressed as:

\begin{equation}
    e_i = g(f([h_i^1, h_i^2, ..., h_i^L]))
\end{equation}

where $e_i$ is the final embedding used for prefix reconstruction.

The contextualization score $C_i$ for token $i$ is then defined as:

\begin{equation}
    C_i = -\log P(w_1, ..., w_{i-1} | e_i)
\end{equation}

where $P(w_1, ..., w_{i-1} | e_i)$ is the probability of the true prefix given the embedding $e_i$.

This methodology allows us to:

\begin{itemize}
    \item Identify which tokens retain the most contextual information.
    \item Analyze how contextualization varies for different types of tokens (e.g., content words vs. function words).
    \item Explore the relationship between contextualization and other properties such as token-level nonlinearity.
    \item Compare contextualization patterns across different model architectures and sizes.
\end{itemize}

\subsection{Examining Intermediate Layers Contribution to Token Prediction}

To track the evolution of token predictions across model's layers, we apply the language model head to intermediate layer representations. Our approach consists of the following steps:
\begin{enumerate}
\item Collect hidden states $h_i^l$ for each token $i$ at each layer $l$.
\item Apply the language model head to obtain token probabilities:
\begin{equation}
p_i^l = \text{softmax}(\text{LM}\text{head}(h_i^l))
\end{equation}
\item Compute prediction error for the next token at each layer:
\begin{equation}
E_i^l = -\log p_i^l[w_{i+1}]
\end{equation}
where $w_{i+1}$ is the true next token.
\end{enumerate}
This analysis shows how prediction accuracy changes across layers, indicating when the model forms its predictions and how confidence evolves. It highlights cases of early correct predictions compared to those requiring full network depth.

\subsection{Visualizing Intermediate Layer Predictions}

In addition to our custom visualization tools, we have implemented the ``Logit Lens'' technique\footnote{\url{https://www.lesswrong.com/posts/AcKRB8wDpdaN6v6ru/interpreting-gpt-the-logit-lens}}. This method provides an intuitive way to visualize the evolution of token predictions across model's layers.

The Logit Lens applies the model's output layer (LM head) to the activations of intermediate layers. This process generates probability distributions over the vocabulary at each layer, offering insight into the model's ``beliefs'' as it processes the input. 

The ``Logit Lens'' suggests that these models primarily ``think in predictive space,'' quickly transforming inputs into predicted outputs and then refining those predictions over the layers. An example of ``Logit Lens'' output in our framework can be found in the Figure~\ref{fig:multilingual_logit_lens}. The developed framework also support multimodal LLM.

\subsection{Intrinsic Dimension of Representations}

To evaluate the complexity and the information content of token representations, we estimate their intrinsic dimensionality using the method proposed by \citet{ID}. This approach examines how the volume of an $n$-dimensional sphere (representing the number of embeddings) scales with dimension $d$.
For each token embedding, we compute:
\begin{equation}
\mu_i = \frac{r_2}{r_1}
\end{equation}
where $r_1$ and $r_2$ are distances to the two nearest neighbors. The intrinsic dimension $d$ is then estimated using:
\begin{equation}
d \approx -\frac{\log(1 - F(\mu))}{\log(\mu)}
\end{equation}
where $F(\mu)$ is the cumulative distribution function of ${\mu_i}$.

\section{Examples and Observations}
% \subsection{Prompt Formats and Induction}

\subsection{The Most Memory Retentive Tokens}

\begin{figure*}[t]
  \centering
   \includegraphics[width=.8\linewidth]{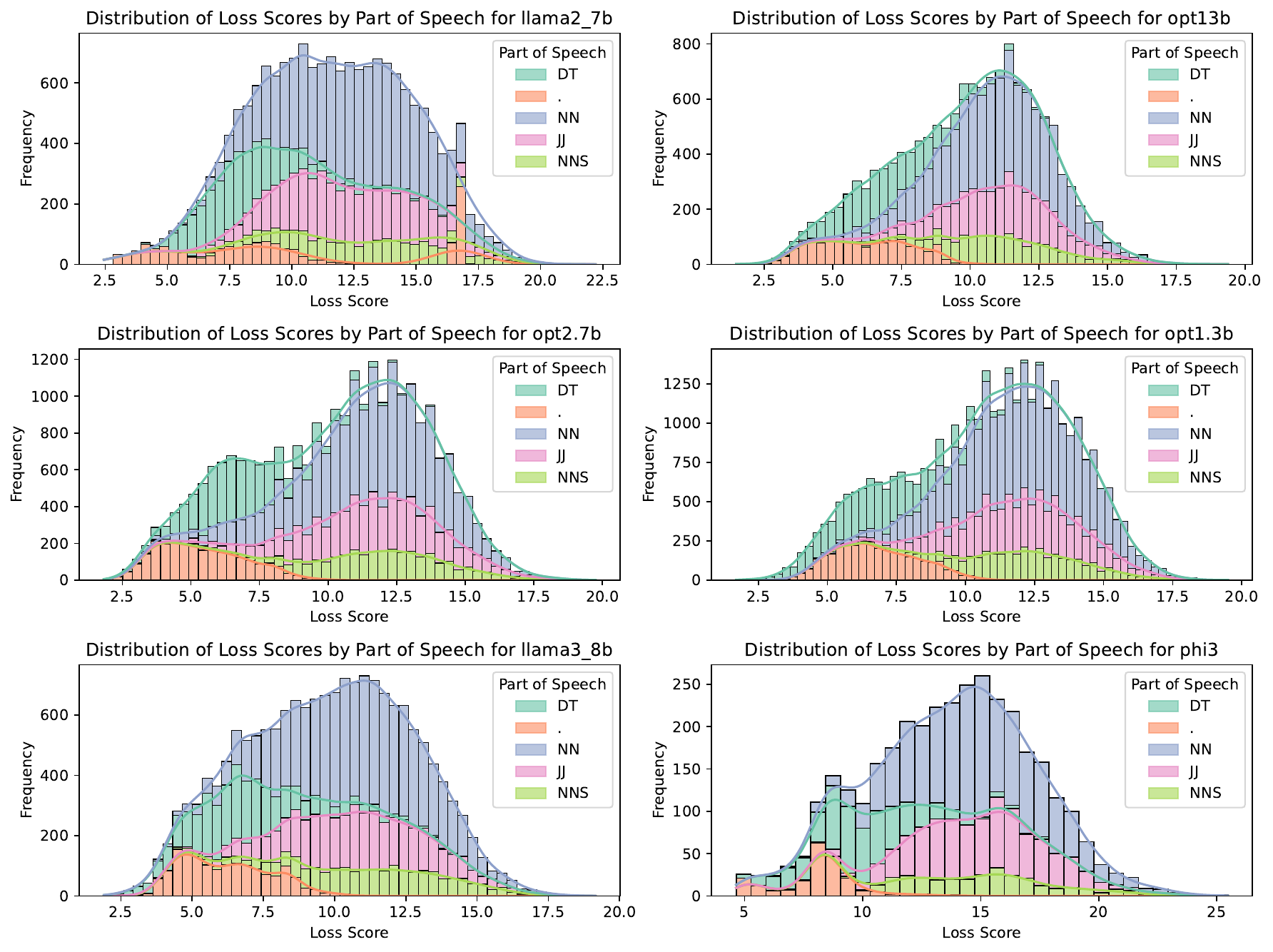}

   \caption{Contextualization score distribution for different parts of speech.}
   \label{fig:histogram}
\end{figure*}

To analyze how different types of tokens retain and encode contextual information, we processed random fragments of Wikipedia articles through our pipeline from Section~\ref{contextuality}, collecting contextualization scores ($C$) for all tokens while preserving information about the original words before tokenization.

Surprisingly, we found out that the tokens that are easiest to use for context (prefix) reconstruction correspond to what are typically considered the least semantically significant elements of language: determiners, prepositions, and punctuation marks. In contrast, nouns and adjectives proved to be the most challenging tokens to reconstruct the prefix.

This counterintuitive finding suggests that language models may use these seemingly less important words as aggregators of memory or overall meaning.

Across all models examined, including different sizes of OPT, Phi, and Llama model families, determiners (DT) and punctuation consistently emerge as the most contextualized tokens with the lowest average reconstruction loss values  $C$.

On the other hand, nouns (NN, NNS) appear universally among the least contextualized tokens, with significantly higher reconstruction loss values $C$. Detailed histograms can be found in the Figure~\ref{fig:histogram}.

\subsection{Examining the Impact of Removing ``Filler'' Tokens}
\label{sec:token_removal_experiments}

\begin{table*}[t]
\centering
\small
\caption{Performance on MMLU and BABILong-4k after partial removal of various token classes, with GPT-4-based removal comparison.}
\label{tab:partial_removal}
\begin{tabular}{lcccccc}
\toprule
\textbf{Model} & 
\textbf{Original} & 
\textbf{No Stop Words} & 
\textbf{No Punctuation} &
\textbf{No Stops \& Punct} &
\textbf{No Articles} &
\textbf{GPT-4 Removal} \\
\midrule
\multicolumn{7}{c}{\textbf{MMLU}} \\
\midrule
Llama-3.2-3B      & 0.398 & 0.347 & 0.391 & 0.342 & 0.386 & 0.377 \\
Mistral-7B-v0.1   & 0.423 & 0.359 & 0.411 & 0.350 & 0.413 & 0.392  \\
meta-llama-3-8B   & 0.430 & 0.365 & 0.419 & 0.351 & 0.415 & 0.403  \\
Qwen2.5-1.5B      & 0.362 & 0.332 & 0.348 & 0.322 & 0.356 & 0.346 \\
\midrule
\multicolumn{7}{c}{\textbf{BABILong 4k}} \\
\midrule
Llama-3.2-3B      & 0.420 & 0.334 & 0.377 & 0.322 & 0.386 & 0.387  \\
Mistral-7B-v0.1   & 0.373 & 0.324 & 0.322 & 0.314 & 0.368 & 0.312  \\
meta-llama-3-8B   & 0.388 & 0.331 & 0.359 & 0.307 & 0.389 & 0.360  \\
Qwen2.5-1.5B      & 0.366 & 0.326 & 0.333 & 0.322 & 0.348 & 0.308  \\
\bottomrule
\end{tabular}
\end{table*}

While our earlier analysis focused on identifying which tokens carry the most contextual information, we also investigated how removing seemingly ``minor'' or ``irrelevant'' tokens affects LLM performance on tasks requiring domain knowledge or extended context. Instead of discarding highly contextualized tokens, we selectively removed punctuation, stopwords, and articles in two distinct modes: (1) a naive, rule-based removal that targets all such tokens, and (2) a more nuanced approach using GPT-4o.

\paragraph{Benchmarks.}
We evaluated these removal strategies on two benchmarks:
\begin{itemize}
    \item \textbf{MMLU} \cite{hendrycks2021}: A widely used multiple-choice benchmark spanning various academic subjects, testing both factual recall and general reasoning. MMLU was evaluated in a zero-shot setting.
    \item \textbf{BABILong-4k} \cite{kuratov2024babilong}: A long-context reasoning benchmark combining facts (from the bAbI dataset \cite{weston2015}) and large amounts of distractor text (from PG19 \cite{rae2019}), where crucial details may be scattered across up to 4k tokens.
\end{itemize}

\paragraph{Removal Conditions.}
We examined several removal strategies:
\begin{enumerate}
    \item \textbf{No Stopwords}: Delete common English function words (e.g., \textit{the, an, and}).
    \item \textbf{No Punctuation}: Remove punctuation marks (commas, periods, quotes, etc.).
    \item \textbf{No Articles}: Remove only English articles (\textit{a, an, the}).
    \item \textbf{No Stopwords \& Punct}: Remove both stopwords and punctuation.
    \item \textbf{GPT-4o Removal}: Prompt GPT-4 to remove only those stopwords or punctuation marks that it deems safe to delete without changing the meaning. 
    Below is the exact system prompt used for GPT-4o when removing tokens:
    \begin{center}
\begin{minipage}{1.\linewidth}
\tiny
\begin{verbatim}
system_message = """
    You are an expert in natural language processing.
    Your task is to remove stop words and punctuation from the user's text
    only when their removal does not alter the meaning of the text.
    Stop words are common words that add little meaning to the text
    (e.g., 'and', 'the', 'in', 'on', 'at', etc.).
    If removing all stop words and punctuation would change the meaning,
    remove only those that contribute the least to the meaning
    while preserving readability.
    Do not rephrase or change the order of words.
    Return only the modified text, without extra commentary.
"""
\end{verbatim}
\end{minipage}
\end{center}
\end{enumerate}

Table~\ref{tab:partial_removal} summarizes the accuracy of several LLMs on MMLU and BABILong-4k under these token-removal schemes. Notably, the deletion of punctuation and basic function words yields a consistent drop in performance. On BABILong-4k, where capturing subtle facts within a large context is crucial, the accuracy losses are especially pronounced.

These results are in line with our earlier findings: LLMs often store key contextual signals in ``filler'' tokens (stopwords, punctuation) that might be seemed unimportant for semantic meaning. Even a carefully controlled removal policy via GPT-4o shows that seemingly trivial tokens play an outsized role in preserving the chain of context — particularly when handling long sequences or academic questions.

\subsection{Correlation Between Nonlinearity and Context Memory}

\begin{figure}
  \centering
   \includegraphics[width=0.9\linewidth]{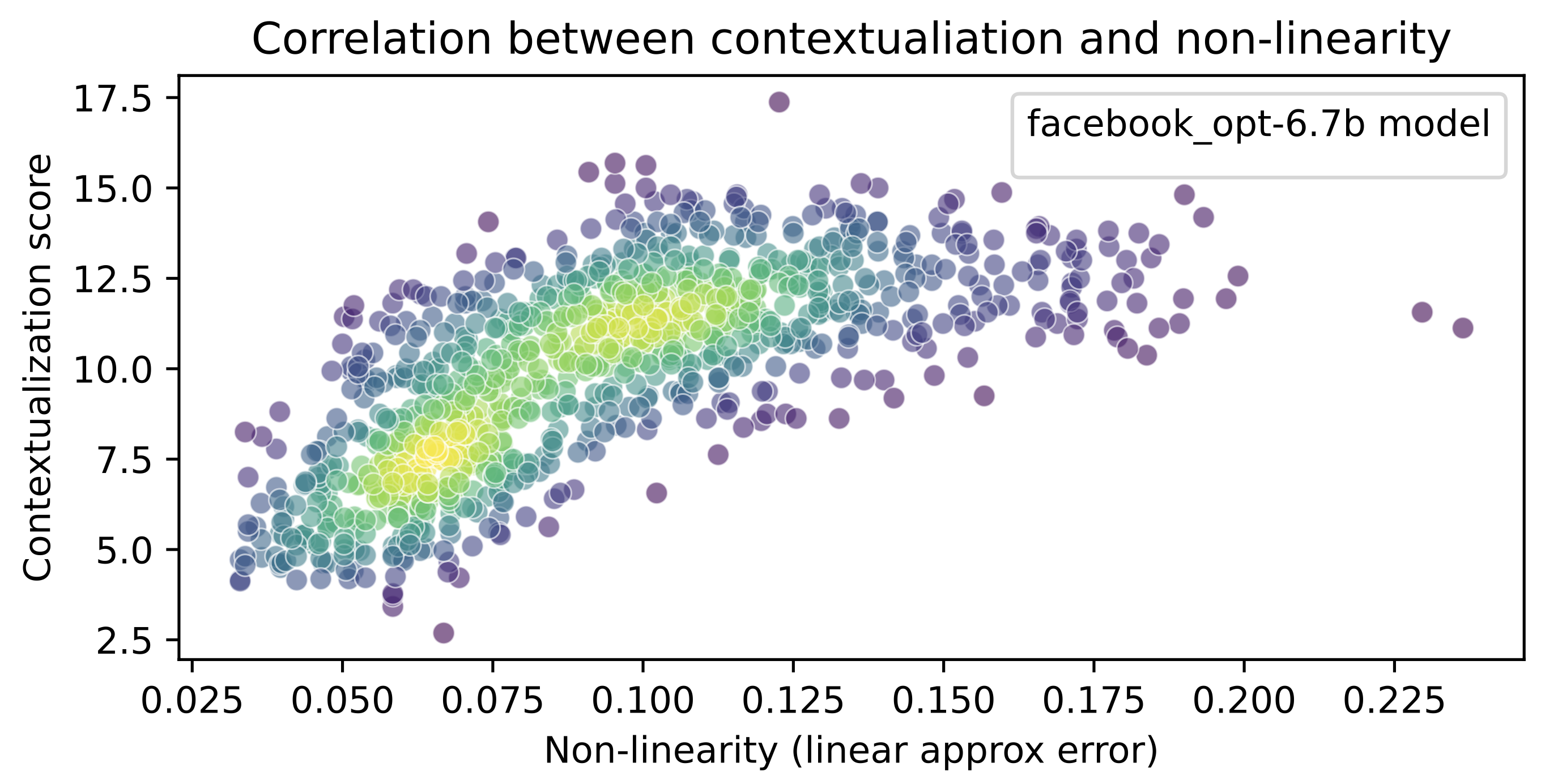}

   \caption{The distribution of Cotextuality $C$ and non-linearity scores for random fragments of text on English Wikipedia articles.}
   \label{fig:correlation}
\end{figure}

\begin{table}[ht]
\small
\caption{Correlation coefficient.}
\centering
\label{tab:correlations}
\begin{tabular}{lc}
\toprule
Model & Corr  \\ % подгоните количество & под количество столбцов!
\midrule
Opt-6.7b                 & 0.482 \\
Opt-1.3b                 & 0.406 \\
Opt-13b                  & 0.359 \\
Opt-2.7b                 & 0.401 \\
Gemma-2-9b               & 0.561 \\
Gemma-2-2b               & 0.515 \\
Llama-3.2-3B             & 0.367 \\
Llama-3 8B               & 0.050 \\
Llama-3 8B Instruct      & 0.328 \\
Llama-2 7Bfp16           & 0.239 \\
Phi-3-mini 128k instruct & 0.410 \\
Qwen2.5 1.5B             & 0.199 \\
Mistral-7B-v0.1          & 0.152 \\
\bottomrule
\small{*p-value less than 0.05 in all measurements}
\end{tabular}
\end{table}

We observed a significant correlation between layer-averaged linearity and contextualization scores for individual tokens. Tokens with high contextualization tend to correspond to the most linear transformations across layers. Figure~\ref{fig:correlation} illustrates this relationship for the OPT-6.7B model, showing the distribution of linearity versus contextualization scores.
This correlation is consistent across different model architectures and sizes, as evidenced by the Pearson correlation coefficients presented in Table~\ref{tab:correlations}.

These findings suggest a potential link between the model's ability to retain contextual information and the linearity of its internal representations.

\subsection{Multilingual Reasoning}

\begin{figure}[h]
  \centering
   \includegraphics[width=0.9\linewidth]{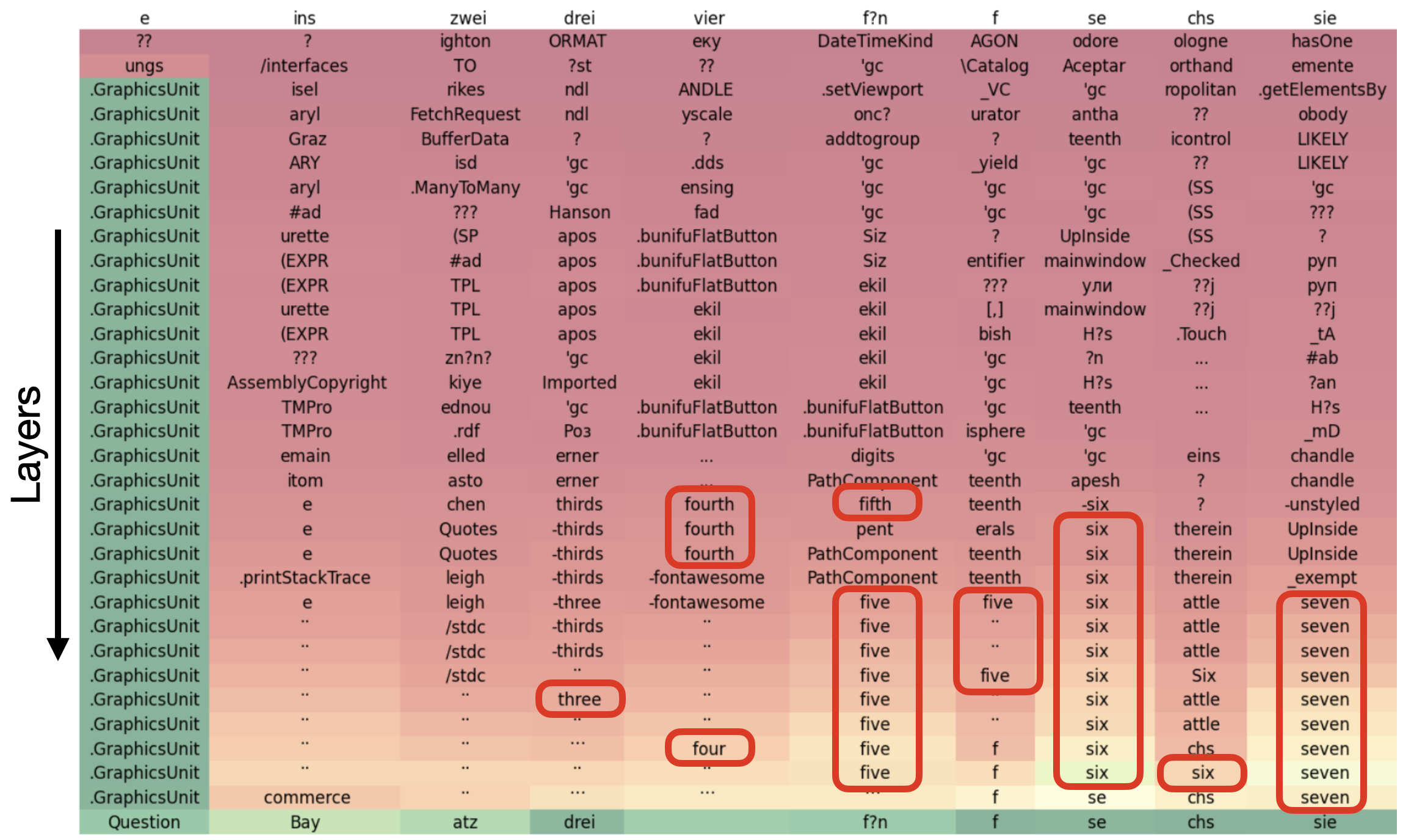}

   \caption{Logit lens visualisation for Llama3-8B. Input text in German: ``eins zwei drei vier fünf sechs sieben,'' which translates into English: ``one two three four five six seven.''}
   \label{fig:multilingual_logit_lens}
\end{figure}

Using the ``Logit Lens'' technique, we studied how language models process non-English input. Our analysis shows that intermediate layer representations predominantly correspond to English tokens, even when the input is in another language.
Figure \ref{fig:multilingual_logit_lens} demonstrates this behavior. The heatmap displays token predictions across layers, with each row representing a layer and each column a token position. The color intensity indicates the model's confidence in its top-1 token prediction. 

Correct non-English tokens corresponding to the translated version of the input gradually emerge in later layers. These observation suggests that the models may perform implicit translation into English before generating the final output.

\section{Conclusion}

In this work, we introduced methods to quantify how Large Language Models (LLMs) encode and store contextual information, revealing the surprising importance of seemingly “minor” tokens — such as determiners, punctuation, and stopwords — in maintaining coherence and context. Our analysis showed a strong correlation between a token’s contextualization level and how linearly one layer’s representation can be mapped onto the next, suggesting a close relationship between model architecture and the retention of contextual cues.

Through empirical evaluations on MMLU and BABILong 4k, we demonstrated that removing high-context tokens — even if they appear trivial — consistently degrades performance. Notably, this effect remains even when a strong language model (GPT-4o) is used to selectively remove only those tokens deemed least relevant. These findings highlight that “filler” tokens can carry critical context, underscoring the need for more refined interpretability approaches.

To facilitate further research in this area, we presented \textbf{LLM-Microscope}, an open-source toolkit that offers: Token-level nonlinearity analysis,  Methods for assessing contextual memory, Visualizations of intermediate layer contributions through an adapted Logit Lens,  Intrinsic dimensionality measurements of internal representations.

\section{Limitations}

\begin{itemize}
\item LM-head application: Using a pre-trained LM-head on intermediate embeddings without fine-tuning may not accurately reflect the actual functionality of these layers.
\item Contextual memory assessment: The adapter-based method's accuracy may be influenced by the adapter's architecture, training data, and optimization process.
\item Generalizability: The results may not be equally applicable to all  model architectures, sizes, or training paradigms.
\end{itemize}

\section{Ethical Statement}

This research aims to improve LLM transparency and interpretability, potentially improving AI safety and reliability. Our tools are designed for analysis only and cannot modify model behavior. We acknowledge the dual-use potential of interpretability research and advocate for responsible use. All experiments were conducted on publicly available pre-trained models without access to personal data or its generation.

This work advances our understanding of LLM internals, contributing to the development of more transparent and reliable natural language processing systems.

\bibliography{acl_latex}

@article{ID,
  author       = {Elena Facco and
                  Maria d'Errico and
                  Alex Rodriguez and
                  Alessandro Laio},
  title        = {Estimating the intrinsic dimension of datasets by a minimal neighborhood
                  information},
  journal      = {CoRR},
  volume       = {abs/1803.06992},
  year         = {2018},
  url          = {http://arxiv.org/abs/1803.06992},
  eprinttype    = {arXiv},
  eprint       = {1803.06992},
  bibsource    = {dblp computer science bibliography, https://dblp.org}
}

@article{tinyStories,
  author       = {Ronen Eldan and
                  Yuanzhi Li},
  title        = {TinyStories: How Small Can Language Models Be and Still Speak Coherent
                  English?},
  journal      = {CoRR},
  volume       = {abs/2305.07759},
  year         = {2023},
  url          = {https://doi.org/10.48550/arXiv.2305.07759},
  doi          = {10.48550/ARXIV.2305.07759},
  eprinttype    = {arXiv},
  eprint       = {2305.07759},
  bibsource    = {dblp computer science bibliography, https://dblp.org}
}

@inproceedings{ettinger-etal-2016-probing,
    title = "Probing for semantic evidence of composition by means of simple classification tasks",
    author = "Ettinger, Allyson  and
      Elgohary, Ahmed  and
      Resnik, Philip",
    booktitle = "Proceedings of the 1st Workshop on Evaluating Vector-Space Representations for {NLP}",
    month = aug,
    year = "2016",
    address = "Berlin, Germany",
    publisher = "Association for Computational Linguistics",
    url = "https://aclanthology.org/W16-2524",
    doi = "10.18653/v1/W16-2524",
    pages = "134--139",
}

@inproceedings{belinkov-etal-2017-neural,
    title = "What do Neural Machine Translation Models Learn about Morphology?",
    author = "Belinkov, Yonatan  and
      Durrani, Nadir  and
      Dalvi, Fahim  and
      Sajjad, Hassan  and
      Glass, James",
    editor = "Barzilay, Regina  and
      Kan, Min-Yen",
    booktitle = "Proceedings of the 55th Annual Meeting of the Association for Computational Linguistics (Volume 1: Long Papers)",
    month = jul,
    year = "2017",
    address = "Vancouver, Canada",
    publisher = "Association for Computational Linguistics",
    url = "https://aclanthology.org/P17-1080",
    doi = "10.18653/v1/P17-1080",
    pages = "861--872",
}

@inproceedings{conneau-etal-2018-cram,
    title = "What you can cram into a single {\$}{\&}!{\#}* vector: Probing sentence embeddings for linguistic properties",
    author = {Conneau, Alexis  and
      Kruszewski, German  and
      Lample, Guillaume  and
      Barrault, Lo{\"\i}c  and
      Baroni, Marco},
    editor = "Gurevych, Iryna  and
      Miyao, Yusuke",
    booktitle = "Proceedings of the 56th Annual Meeting of the Association for Computational Linguistics (Volume 1: Long Papers)",
    month = jul,
    year = "2018",
    address = "Melbourne, Australia",
    publisher = "Association for Computational Linguistics",
    url = "https://aclanthology.org/P18-1198",
    doi = "10.18653/v1/P18-1198",
    pages = "2126--2136",
}

@article{belinkov2022probing,
  title={Probing classifiers: Promises, shortcomings, and advances},
  author={Belinkov, Yonatan},
  journal={Computational Linguistics},
  volume={48},
  number={1},
  pages={207--219},
  year={2022},
  publisher={MIT Press One Broadway, 12th Floor, Cambridge, Massachusetts 02142, USA~…}
}

@article{calderon2024behalf,
  title={On Behalf of the Stakeholders: Trends in NLP Model Interpretability in the Era of LLMs},
  author={Calderon, Nitay and Reichart, Roi},
  journal={arXiv preprint arXiv:2407.19200},
  year={2024}
}

@inproceedings{dai-etal-2022-knowledge,
    title = "Knowledge Neurons in Pretrained Transformers",
    author = "Dai, Damai  and
      Dong, Li  and
      Hao, Yaru  and
      Sui, Zhifang  and
      Chang, Baobao  and
      Wei, Furu",
    editor = "Muresan, Smaranda  and
      Nakov, Preslav  and
      Villavicencio, Aline",
    booktitle = "Proceedings of the 60th Annual Meeting of the Association for Computational Linguistics (Volume 1: Long Papers)",
    month = may,
    year = "2022",
    address = "Dublin, Ireland",
    publisher = "Association for Computational Linguistics",
    url = "https://aclanthology.org/2022.acl-long.581",
    doi = "10.18653/v1/2022.acl-long.581",
    pages = "8493--8502",
}

@inproceedings{conmy2023,
 author = {Conmy, Arthur and Mavor-Parker, Augustine and Lynch, Aengus and Heimersheim, Stefan and Garriga-Alonso, Adri\`{a}},
 booktitle = {Advances in Neural Information Processing Systems},
 editor = {A. Oh and T. Naumann and A. Globerson and K. Saenko and M. Hardt and S. Levine},
 pages = {16318--16352},
 publisher = {Curran Associates, Inc.},
 title = {Towards Automated Circuit Discovery for Mechanistic Interpretability},
 url = {https://proceedings.neurips.cc/paper_files/paper/2023/file/34e1dbe95d34d7ebaf99b9bcaeb5b2be-Paper-Conference.pdf},
 volume = {36},
 year = {2023}
}

@article{meng2022locating,
  title={Locating and Editing Factual Associations in {GPT}},
  author={Kevin Meng and David Bau and Alex Andonian and Yonatan Belinkov},
  journal={Advances in Neural Information Processing Systems},
  volume={36},
  year={2022},
  note={arXiv:2202.05262}
}

@misc{wang2022interpretabilitywildcircuitindirect,
      title={Interpretability in the Wild: a Circuit for Indirect Object Identification in GPT-2 small}, 
      author={Kevin Wang and Alexandre Variengien and Arthur Conmy and Buck Shlegeris and Jacob Steinhardt},
      year={2022},
      eprint={2211.00593},
      archivePrefix={arXiv},
      primaryClass={cs.LG},
      url={https://arxiv.org/abs/2211.00593}, 
}

@misc{cunningham2023sparseautoencodershighlyinterpretable,
      title={Sparse Autoencoders Find Highly Interpretable Features in Language Models}, 
      author={Hoagy Cunningham and Aidan Ewart and Logan Riggs and Robert Huben and Lee Sharkey},
      year={2023},
      eprint={2309.08600},
      archivePrefix={arXiv},
      primaryClass={cs.LG},
      url={https://arxiv.org/abs/2309.08600}, 
}

@inproceedings{yu-etal-2023-characterizing,
    title = "Characterizing Mechanisms for Factual Recall in Language Models",
    author = "Yu, Qinan  and
      Merullo, Jack  and
      Pavlick, Ellie",
    editor = "Bouamor, Houda  and
      Pino, Juan  and
      Bali, Kalika",
    booktitle = "Proceedings of the 2023 Conference on Empirical Methods in Natural Language Processing",
    month = dec,
    year = "2023",
    address = "Singapore",
    publisher = "Association for Computational Linguistics",
    url = "https://aclanthology.org/2023.emnlp-main.615",
    doi = "10.18653/v1/2023.emnlp-main.615",
    pages = "9924--9959",
}

@article{gao2024,
  author       = {Leo Gao and
                  Tom Dupr{\'{e}} la Tour and
                  Henk Tillman and
                  Gabriel Goh and
                  Rajan Troll and
                  Alec Radford and
                  Ilya Sutskever and
                  Jan Leike and
                  Jeffrey Wu},
  title        = {Scaling and evaluating sparse autoencoders},
  journal      = {CoRR},
  volume       = {abs/2406.04093},
  year         = {2024},
  url          = {https://doi.org/10.48550/arXiv.2406.04093},
  doi          = {10.48550/ARXIV.2406.04093},
  eprinttype    = {arXiv},
  eprint       = {2406.04093},
  bibsource    = {dblp computer science bibliography, https://dblp.org}
}

@article{templeton2024scaling,
       title={Scaling Monosemanticity: Extracting Interpretable Features from Claude 3 Sonnet},
       author={Templeton, Adly and Conerly, Tom and Marcus, Jonathan and Lindsey, Jack and Bricken, Trenton and Chen, Brian and Pearce, Adam and Citro, Craig and Ameisen, Emmanuel and Jones, Andy and Cunningham, Hoagy and Turner, Nicholas L and McDougall, Callum and MacDiarmid, Monte and Freeman, C. Daniel and Sumers, Theodore R. and Rees, Edward and Batson, Joshua and Jermyn, Adam and Carter, Shan and Olah, Chris and Henighan, Tom},
       year={2024},
       journal={Transformer Circuits Thread},
       url={https://transformer-circuits.pub/2024/scaling-monosemanticity/index.html}
    }

@misc{nostalgebraist2020,
  author = {Nostalgebraist},
  title = {{interpreting GPT: the logit lens}},
  year = {2020},
  howpublished = {\url{https://www.alignmentforum.org/posts/AcKRB8wDpdaN6v6ru/interpreting-gpt-the-logit-lens}}
}

@inproceedings{xu-etal-2021-probing,
    title = "Probing Word Translations in the Transformer and Trading Decoder for Encoder Layers",
    author = "Xu, Hongfei  and
      van Genabith, Josef  and
      Liu, Qiuhui  and
      Xiong, Deyi",
    editor = "Toutanova, Kristina  and
      Rumshisky, Anna  and
      Zettlemoyer, Luke  and
      Hakkani-Tur, Dilek  and
      Beltagy, Iz  and
      Bethard, Steven  and
      Cotterell, Ryan  and
      Chakraborty, Tanmoy  and
      Zhou, Yichao",
    booktitle = "Proceedings of the 2021 Conference of the North American Chapter of the Association for Computational Linguistics: Human Language Technologies",
    month = jun,
    year = "2021",
    address = "Online",
    publisher = "Association for Computational Linguistics",
    url = "https://aclanthology.org/2021.naacl-main.7",
    doi = "10.18653/v1/2021.naacl-main.7",
    pages = "74--85",
}

@misc{belrose2023eliciting,
      title={Eliciting Latent Predictions from Transformers with the Tuned Lens}, 
      author={Nora Belrose and Zach Furman and Logan Smith and Danny Halawi and Igor Ostrovsky and Lev McKinney and Stella Biderman and Jacob Steinhardt},
      year={2023},
      eprint={2303.08112},
      archivePrefix={arXiv},
      primaryClass={cs.LG}
}

@misc{din2023jump,
      title={Jump to Conclusions: Short-Cutting Transformers With Linear Transformations}, 
      author={Alexander Yom Din and Taelin Karidi and Leshem Choshen and Mor Geva},
      year={2023},
      eprint={2303.09435},
      archivePrefix={arXiv},
      primaryClass={cs.CL}
}

@misc{anthropic2021,
      title={A Mathematical Framework for Transformer Circuits}, 
      author={Nelson Elhage and Neel Nanda and Catherine Olsson and Tom Henighan† and Nicholas Joseph† and Ben Mann† and Amanda Askell and Yuntao Bai and Anna Chen and Tom Conerly and Nova DasSarma and Dawn Drain and Deep Ganguli and Zac Hatfield-Dodds and Danny Hernandez and Andy Jones and Jackson Kernion and Liane Lovitt and Kamal Ndousse and Dario Amodei and Tom Brown and Jack Clark and Jared Kaplan and Sam McCandlish and Chris Olah‡},
      year={2021},
      eprint={https://transformer-circuits.pub/2021/framework/index.html},
}

@article{olsson2022context,
   title={In-context Learning and Induction Heads},
   author={Olsson, Catherine and Elhage, Nelson and Nanda, Neel and Joseph, Nicholas and DasSarma, Nova and Henighan, Tom and Mann, Ben and Askell, Amanda and Bai, Yuntao and Chen, Anna and Conerly, Tom and Drain, Dawn and Ganguli, Deep and Hatfield-Dodds, Zac and Hernandez, Danny and Johnston, Scott and Jones, Andy and Kernion, Jackson and Lovitt, Liane and Ndousse, Kamal and Amodei, Dario and Brown, Tom and Clark, Jack and Kaplan, Jared and McCandlish, Sam and Olah, Chris},
   year={2022},
   journal={Transformer Circuits Thread},
   note={https://transformer-circuits.pub/2022/in-context-learning-and-induction-heads/index.html}
}

@inproceedings{razzhigaev-etal-2024-shape,
    title = "The Shape of Learning: Anisotropy and Intrinsic Dimensions in Transformer-Based Models",
    author = "Razzhigaev, Anton  and
      Mikhalchuk, Matvey  and
      Goncharova, Elizaveta  and
      Oseledets, Ivan  and
      Dimitrov, Denis  and
      Kuznetsov, Andrey",
    editor = "Graham, Yvette  and
      Purver, Matthew",
    booktitle = "Findings of the Association for Computational Linguistics: EACL 2024",
    month = mar,
    year = "2024",
    address = "St. Julian{'}s, Malta",
    publisher = "Association for Computational Linguistics",
    url = "https://aclanthology.org/2024.findings-eacl.58",
    pages = "868--874",
}

@misc{razzhigaev2024transformersecretlylinear,
      title={Your Transformer is Secretly Linear}, 
      author={Anton Razzhigaev and Matvey Mikhalchuk and Elizaveta Goncharova and Nikolai Gerasimenko and Ivan Oseledets and Denis Dimitrov and Andrey Kuznetsov},
      year={2024},
      eprint={2405.12250},
      archivePrefix={arXiv},
      primaryClass={cs.LG},
      url={https://arxiv.org/abs/2405.12250}, 
}

@misc{wan2024informationleakageembeddinglarge,
      title={Information Leakage from Embedding in Large Language Models}, 
      author={Zhipeng Wan and Anda Cheng and Yinggui Wang and Lei Wang},
      year={2024},
      eprint={2405.11916},
      archivePrefix={arXiv},
      primaryClass={cs.LG},
      url={https://arxiv.org/abs/2405.11916}, 
}

@inproceedings{li-etal-2023-sentence,
    title = "Sentence Embedding Leaks More Information than You Expect: Generative Embedding Inversion Attack to Recover the Whole Sentence",
    author = "Li, Haoran  and
      Xu, Mingshi  and
      Song, Yangqiu",
    editor = "Rogers, Anna  and
      Boyd-Graber, Jordan  and
      Okazaki, Naoaki",
    booktitle = "Findings of the Association for Computational Linguistics: ACL 2023",
    month = jul,
    year = "2023",
    address = "Toronto, Canada",
    publisher = "Association for Computational Linguistics",
    url = "https://aclanthology.org/2023.findings-acl.881",
    doi = "10.18653/v1/2023.findings-acl.881",
    pages = "14022--14040",
}

@misc{yao2024knowledgecircuitspretrainedtransformers,
      title={Knowledge Circuits in Pretrained Transformers}, 
      author={Yunzhi Yao and Ningyu Zhang and Zekun Xi and Mengru Wang and Ziwen Xu and Shumin Deng and Huajun Chen},
      year={2024},
      eprint={2405.17969},
      archivePrefix={arXiv},
      primaryClass={cs.CL},
      url={https://arxiv.org/abs/2405.17969}, 
}

@article{radford2019language,
  title={Language Models are Unsupervised Multitask Learners},
  author={Radford, Alec and Wu, Jeff and Child, Rewon and Luan, David and Amodei, Dario and Sutskever, Ilya},
  year={2019}
}

@misc{zhang2024tinyllama,
      title={TinyLlama: An Open-Source Small Language Model}, 
      author={Peiyuan Zhang and Guangtao Zeng and Tianduo Wang and Wei Lu},
      year={2024},
      eprint={2401.02385},
      archivePrefix={arXiv},
      primaryClass={cs.CL},
      url={https://arxiv.org/abs/2401.02385}, 
}

@misc{hendrycks2021,
      title={Measuring Massive Multitask Language Understanding}, 
      author={Dan Hendrycks and Collin Burns and Steven Basart and Andy Zou and Mantas Mazeika and Dawn Song and Jacob Steinhardt},
      year={2021},
      eprint={2009.03300},
      archivePrefix={arXiv},
      primaryClass={cs.CY},
      url={https://arxiv.org/abs/2009.03300}, 
}

@misc{kuratov2024babilong,
      title={BABILong: Testing the Limits of LLMs with Long Context Reasoning-in-a-Haystack}, 
      author={Yuri Kuratov and Aydar Bulatov and Petr Anokhin and Ivan Rodkin and Dmitry Sorokin and Artyom Sorokin and Mikhail Burtsev},
      year={2024},
      eprint={2406.10149},
      archivePrefix={arXiv},
      primaryClass={id='cs.CL' full_name='Computation and Language' is_active=True alt_name='cmp-lg' in_archive='cs' is_general=False description='Covers natural language processing. Roughly includes material in ACM Subject Class I.2.7. Note that work on artificial languages (programming languages, logics, formal systems) that does not explicitly address natural-language issues broadly construed (natural-language processing, computational linguistics, speech, text retrieval, etc.) is not appropriate for this area.'}
}

@misc{weston2015,
      title={Towards AI-Complete Question Answering: A Set of Prerequisite Toy Tasks}, 
      author={Jason Weston and Antoine Bordes and Sumit Chopra and Alexander M. Rush and Bart van Merriënboer and Armand Joulin and Tomas Mikolov},
      year={2015},
      eprint={1502.05698},
      archivePrefix={arXiv},
      primaryClass={cs.AI},
      url={https://arxiv.org/abs/1502.05698}, 
}

@misc{rae2019,
      title={Compressive Transformers for Long-Range Sequence Modelling}, 
      author={Jack W. Rae and Anna Potapenko and Siddhant M. Jayakumar and Timothy P. Lillicrap},
      year={2019},
      eprint={1911.05507},
      archivePrefix={arXiv},
      primaryClass={cs.LG},
      url={https://arxiv.org/abs/1911.05507}, 
}

\end{document}